\documentclass{article}

\usepackage{arxiv}

\usepackage[utf8]{inputenc} 
\usepackage[T1]{fontenc}    
\usepackage{booktabs}       
\usepackage{amsfonts}       
\usepackage{nicefrac}       
\usepackage{microtype}      
\usepackage{lipsum}		
\usepackage{natbib}
\usepackage{doi}
\usepackage{hyperref}
\usepackage{url}
\usepackage{graphicx}
\usepackage{amssymb}
\usepackage{amsmath}

\title{Preference Packing: Efficient Preference Optimization for Large Language Models}

\author{Jaekyung Cho\thanks{This publication was written prior to the author joining Amazon.} \\
AWS GenAI Innovation Center\\
\texttt{jackcho@amazon.com}
}

\renewcommand{\shorttitle}{\textit{arXiv} Template}

\hypersetup{
pdftitle={A template for the arxiv style},
pdfsubject={q-bio.NC, q-bio.QM},
pdfauthor={Jaekyung Cho},
pdfkeywords={LLM, Preference optimization, DPO},
}

\begin{document}
\maketitle

\begin{abstract}
Resource-efficient training optimization techniques are becoming increasingly important as the size of large language models (LLMs) continues to grow. In particular, batch packing is commonly used in pre-training and supervised fine-tuning to achieve resource-efficient training. We propose \textbf{preference packing}, a method to enhance resource efficiency in training techniques that use data with different responses for the same input prompt, such as reward models or Direct Preference Optimization (DPO). Preference packing improves resource efficiency by reducing the attention operations for duplicate input prompts and decreasing KV cache memory usage. We conducted experiments on text-only datasets and image-included datasets and achieved at least 37\% reduction in training time. Notably, this method can be applied alongside existing optimization techniques such as batch sorting, resulting in a $3.22\times$ speedup.
\end{abstract}

\section{Introduction}
The discovery of Large Language Model (LLM) scaling laws has played a crucial role in rapidly advancing LLM performance \cite{kaplan2020scaling}. Increasing model parameter size and dataset size has significantly improved performance but has also required massive GPU clusters and extensive computation time. Given the immense resource demands, efficiently utilizing time and computational resources has become a critical challenge in LLM development.

Various resource-efficient training techniques have been used for pre-training and supervised fine-tuning. Batch packing was among the first methods studied, where datasets with different sequence lengths are concatenated within a batch, and training is performed by adjusting the causal mask instead of using padding \cite{chung2024scaling}. Additionally, batch sorting methods have been explored to improve efficiency by grouping sequences of similar lengths into the same batch, reducing resource waste caused by padding \cite{bai2024longalign}.

Unlike traditional pre-training and supervised fine-tuning (SFT) using next-token prediction loss, alignment tuning methods have emerged to learn human preferences. Among them, reward model training \cite{liu2020learning,ouyang2022training} and Direct Preference Optimization (DPO) \cite{rafailov2024direct} utilize data where different responses to the same input prompt are ranked. However, the input prompt, which constitutes the majority of the data, leads to redundant computations and duplicated memory usage. Until now, there has been no method to handle these issues in a resource-efficient manner.

Therefore, we propose \textbf{preference packing} for efficient alignment tuning. Instead of grouping multiple responses to the same input prompt into a batch, we sequentially pack the responses. We then construct a causal attention mask for the packed sequence, ensuring that computations and KV memory caching for the repeated input prompt occur only once. 

In this work, we propose a resource-efficient technique for alignment tuning, demonstrating its computational and memory advantages on real-world preference datasets. Furthermore, we experimentally validate that our method remains effective in distributed training settings of large-scale LLMs with up to 72B parameters.


\section{Related Work}
\label{related_works}

\subsection{Preference optimization}
\label{dpo_and_rm}
Reinforcement Learning from Human Feedback (RLHF) methods use data where multiple responses exist for the same input prompt, ranking them based on preference \cite{liu2020learning,ouyang2022training}. The reward model (RM) used in RLHF leverages the Bradley-Terry model to learn preferences and predict a scalar reward value for a given token sequence. We notate $x$ as a single input prompt, $y_w$ as a preferable response, and $y_l$ as a relatively not preferable response. The loss function of the RM $r_\theta$ is as following:
$$
\mathcal{L}_\theta = \mathbb{E}_{(x, y_w, y_l) \sim \mathcal{D}} \left[ \log \sigma \left( r_\theta(x, y_w) - r_\theta(x, y_l) \right) \right]
$$

Direct Preference Optimization (DPO)\cite{rafailov2024direct}, which is a simpler alternative to RLHF, directly trains the LLM using the same preference data that was used to train the reward model. It employs a loss function same that is identical in form to that of the reward model, but the key difference is that it leverages the probability ratio between the learning model and a reference model as a reward. This allows the model to be trained to increase the probability of the preferred response $y_w$ and decrease that of the less preferred response $y_l$ for the same input $x$.
$$
r_\theta^\text{DPO} = \beta\frac{\pi_\theta(x,y)}{\pi_{\text{ref}}(x,y)}
$$
Although preference datasets usually contain pairwise chosen and rejected responses for a single input prompt as represented above, they can be extended to multiple ranked responses for a single input prompt \cite{ouyang2022training}.

\subsection{Training resource efficiency}
\label{training_resource_efficiency}
Raffel et al. \cite{raffel2020exploring} first proposed a packing method to maximize resource utilization without relying on padding within a batch. When combining sequences of varying lengths into a single batch, padding is required to match the longest sequence, which reduces training efficiency. To address this, their method concatenates individual sequences in a sequence-wise manner up to a maximum sequence length before batching. This approach minimizes padding and enables maximum utilization of resources at each update step.

In distributed learning with data parallelism, when the lengths of input samples vary significantly, GPUs processing shorter inputs must wait for the GPU handling the longest input to finish. This idle time leads to a significant drop in training efficiency. To address this issue, Bai et al. \cite{bai2024longalign} proposed a method called batch sorting. By grouping samples of similar lengths into the same batch, they minimized idle time, and experimentally demonstrated that sorting does not negatively affect model performance.



\section{Methods}
\label{methods}

\subsection{Symbols}
\label{symbols}
We assume $\mathcal{D}=\{d_1,d_2,\dots,d_N\}$ denotes the preference dataset where $N$ is the size of the dataset. $d_i=\{(x_i,y_i^1),(x_i,y_i^2),\dots,(x_i,y_i^K)\}, K\geq2$ denotes a preference data point where $x_i$ is a single input prompt and $\{y_i^k\}_{k=1}^K$ are responses from multiple sources. For convenience, we assume that $y_i^\alpha$ is more human-preferable than $y_i^\beta$ for $\alpha > \beta$.

Batching in deep learning normally means grouping $M$ data points into a batch and feeding it into the model to maximize GPU parallel computing performance. Since a preference data consists of multiple sequences, the effective batch size increases by a factor of the number of responses $K$.
\begin{gather*}
    \mathcal{B} = \{b_0,b_1,\dots,b_{N/M-1}\} \\
    b_i=\{d_{iM},\dots,d_{(i+1)M-1}\}=\{(x_{iM},y_{iM}^1),\dots,(x_{(i+1)M-1},y_{(i+1)M-1}^K)\}
\end{gather*}

While the model processes batch $b_i$, the input sequences from $x_{Mi}$ to $x_{M(i+1)-1}$ are repeatedly computed $K$ times. Additionally, the attention memory for these sequences is also redundantly occupied $K$ times.

\subsection{Preference packing}
\label{preference_packing}
\begin{figure}[t]
\centering
\includegraphics[width=0.6\linewidth]{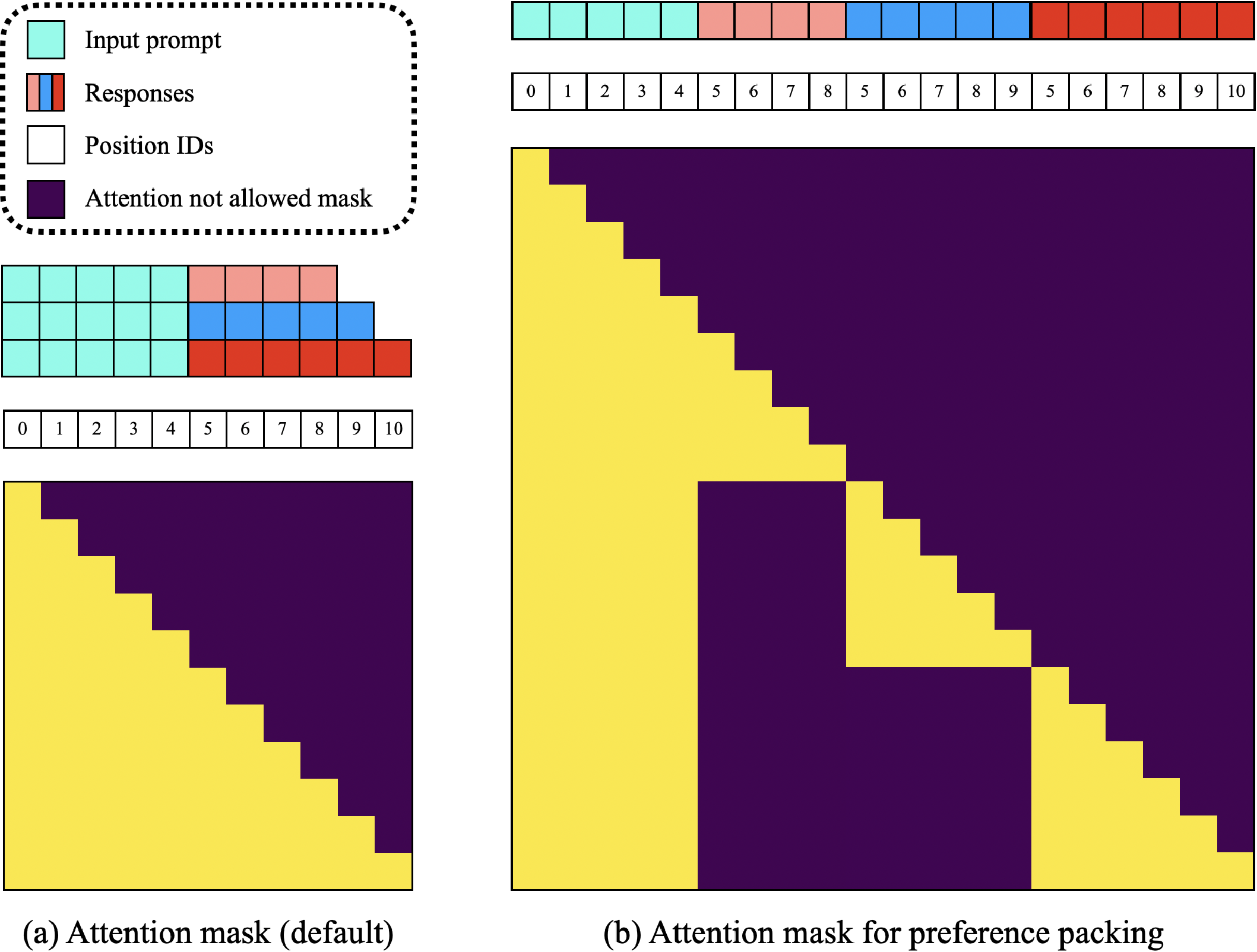}
\caption{Difference in data sequence format, position IDs, and attention masks. (a) By default, a single preference data sample is processed in a batch-wise manner, where input prompt unnecessarily repeated. (b) In preference packing, multiple preference responses are concatenated in a sequence-wise manner into a single sequence following the shared input prompt. Position IDs and attention masks are modified accordingly to reflect the new structure.}
\label{fig_attn_mask}
\end{figure}

To prevent inefficient computation and redundant memory usage, we restructure a single preference data into a single sequence so that the input prompt is processed only once.
$$
d_i=\{(x_i,y_i^1,y_i^2,\dots,y_i^K)\}
$$
This single sequence uses a new preference packing attention mask and position IDs to produce identical computational results as the original format. Figure \ref{fig_attn_mask} shows the details.
The new preference packing attention mask ensures that each response attends only to the input prompt and not to the other responses. Additionally, position IDs help positional embeddings retain their original positions.

\subsubsection{Theoretical performance}
\label{theoretical_preformance}
The preference packing method may appear to require more attention computations and memory as the total sequence length increases; however, its efficiency depends on the length ratio between the input prompt and responses, as well as the number of responses.

Let $l_{in}$ be the length of the input prompt, $l_{resp}^k$ the length of the $k$-th response, and $K$ the number of responses. Note that both the attention computation and memory usage are proportional to the square of the sequence length. 
In practice, sequences are padded to the maximum response length within the batch. Therefore, the effective sequence length becomes
$l_{original}=l_{in}+\max_k l_{resp}^k$. Assuming identical model architecture and attention implementation, the attention computation is proportional to the square of the sequence length. Ignoring constant factors, we have
\[
C_{\text{original}} \propto K \, l_{\text{original}}^2.
\]
For preference packing, the total sequence length is $l_{\text{pp}} = l_{\text{in}} + \sum_{k=1}^K l_{\text{resp}}^k$, and the corresponding attention cost satisfies
\[
C_{\text{pp}} \propto l_{\text{pp}}^2.
\]
In formula, the ratio of attention costs is as follows,
$$
\frac{C_{pp}}{C_{original}} = \frac{(l_{in}+\sum l_{resp}^{k})^2}{K\times(l_{in}+l_{resp}^{max})^2}
$$
For preference packing to yield a strictly lower attention cost, this ratio must be less than 1. Therefore, preference packing is more efficient when the following condition is satisfied.
\begin{equation}
    l_{in}+\sum l_{resp}^{k} < \sqrt{K}\times(l_{in}+l_{resp}^{max})
    \label{eq:main}
\end{equation}
This means that preference packing becomes more efficient when the length differences between responses are larger.

When using Flash Attention \cite{dao2022flashattention}, attention memory has a linear relationship with sequence length. Therefore, assuming Flash Attention is used, the attention memory ratio is as follows.
\begin{equation*}
\frac{C_{pp}}{C_{original}}\vert_{memory} = \frac{l_{in}+\sum l_{resp}^{k}}{K\times(l_{in}+l_{resp}^{max})} < 1
\end{equation*}
This shows that the attention memory ratio is at most 1 and becomes strictly smaller when response lengths vary. Therefore, using Flash Attention together with preference packing guarantees an improvement in memory efficiency.

\section{Experimental Setups}
\label{experimental_setups}

\subsection{Training datasets}
\label{training_datasets}

We verified the effect of preference packing on three publicly available preference datasets. 


\textbf{Orca-DPO-pair}.
The Orca-DPO-pair dataset is one of the earlier and widely used DPO datasets and consists of approximately 12.7K pairs. It is a composite of data from many different categories sampled from OpenOrca \cite{OpenOrca}. GPT-3.5 and GPT-4 APIs are used to construct response pairs.

\textbf{Distilabel Capybara}
The Capybara Preference dataset \cite{distilabel-argilla-2024} consists of 7.56K data pairs and is characterized by its multi-turn conversation input prompts. Although multi-turn conversation is one of the key abilities for LLMs, few open-source preference datasets are available.

\textbf{RLHF-V}
The RLHF-V dataset consists of 5.7K visual question answering pairs. The RLHF process for VLMs is identical to that of LLMs, and images are processed as sequences similarly to text tokens.

Note that all three datasets consist of pairs. Even though \cite{ouyang2022training} and \cite{dubey2024llama} mentioned the use of preference datasets with more than three responses, almost all open-source datasets are composed of pairs.

\subsection{Training Models}
\label{training_models}
We used the open-source model Llama-3.2-1B-Instruct \cite{dubey2024llama} and llava-qwen-0.5b-hf \cite{li2024llavanextinterleavetacklingmultiimagevideo} for performance comparison. 

We compared the training time and peak memory ratio with and without preference packing in a very simple setting: a single GPU using LoRA ($r=1$). The experiments do not require complex configurations because the cost efficiency of preference packing does not strongly correlate with model size or the number of trainable parameters.


\section{Results and Analysis}
\label{results_and_analysis}

\subsection{Resource efficiency}
\label{resource_efficiency}

Table \ref{tab:pref-packing-efficiency} shows the reduction in total training time and peak memory usage for each dataset. Since all datasets consist of pairwise responses ($K = 2$), Eq.~\ref{eq:main} indicates that preference packing becomes more efficient when responses are relatively short compared to the input. As shown in Table \ref{tab:pref-packing-efficiency}, this condition is largely satisfied in the evaluated datasets, which aligns with the observed reductions in training time.

\begin{table}
\caption{Memory and training time reduction across different datasets}
    \centering
    \begin{tabular}{rccc}
    \toprule
            \textbf{Dataset}&  \textbf{Orca}&  \textbf{Capybara}&\textbf{RLAIF-V}\\
    \midrule
   input len.& 228.2& 720.6&599.8\\
  max resp. len.& 251.0& 468.8&109.6\\
  min resp. len.& 129.6& 330.7&89.7\\
  \midrule
            peak memory $\downarrow$&  0.671&   0.802&0.635\\
            time $\downarrow$&  0.801&   0.780&0.798\\
\midrule
 \textbf{effective time $\downarrow$}& \textbf{0.537}& \textbf{0.626}&\textbf{0.507}\\
 \bottomrule
    \end{tabular}
    \label{tab:pref-packing-efficiency}
\end{table}
In the case of the RLAIF-V dataset, the image included in the input is converted to thousands of tokens, significantly increasing the length of the input prompt. Meanwhile, the response consists only of text, making it relatively short. This indicates that preference packing is particularly effective for such data.

Note that the training time reduction we measured assumes the same batch size is used. However, the reduction in peak memory usage with preference packing allows for using larger batch sizes, which can further reduce the total time required to train on the entire dataset. Therefore, assuming full utilization of available GPU resources, the total training time can be reduced to approximately $0.507\times$ of the original training time.

\subsection{Scalability}
\label{usage-independent}
We applied preference packing in DPO training for Qwen2.5-72B-Instruct \cite{qwen2.5}. The model was trained on the Tulu3 preference dataset \cite{lambert2024tulu3}, which contains over 300K samples of varying lengths, all under 16K tokens. Unlike the experiments in Section \ref{resource_efficiency}, the training was conducted in a multi-node distributed environment using FSDP, with all model parameters fully trainable. 

Since we used the distributed training setup, batch sorting can be applied to improve training efficiency. Furthermore, batch sorting can also be applied to preference-packed sequences in the same manner. We measured the training efficiency improvements from preference packing and batch sorting both separately and in combination. The results are shown in Table \ref{tab:large-scale-model-efficiency}.

\begin{table}
    \caption{Increased training efficiency on large-scale model using FSDP}
    \centering
    \begin{tabular}{rc}
    \toprule
         &  \textbf{samples/sec}\\
    \midrule
 vanilla DPO& 1\\
         w/ preference packing&  1.53\\
         w/ batch sorting&  2.62\\
         \midrule
 \textbf{w/ both}& \textbf{3.22}\\
 \bottomrule
    \end{tabular}
    \label{tab:large-scale-model-efficiency}
\end{table}

Preference packing and batch sorting yielded speedups of 1.53$\times$ and 2.62$\times$, respectively. When batch sorting was applied based on sequence length in conjunction with preference packing, the training speedup increased to 3.22$\times$. This corresponds to approximately 23\% additional speedup over using batch sorting alone.


\section*{Conclusion}
\label{conclusion}
In this paper, we introduce a novel method called preference packing to enable more efficient resource utilization when training LLMs with preference data. This approach eliminates redundant computations over repeated input sequences in preference optimization, where multiple outputs share the same input. We provide a theoretical bound on the efficiency of preference packing and empirically demonstrate that it can reduce memory usage by up to 80\% and training time by up to 50\%. Furthermore, we validate its practical effectiveness by accelerating the training of a large-scale LLM by up to 3.22 times when combined with batch sorting.

\section*{Limitations}
\label{limitations}
The proposed preference packing method can actually increase training cost when the responses are longer than the input prompts. Especially in the case of reasoning models, it is difficult to reduce training cost using this method because responses tend to be extremely long.

\bibliographystyle{plain}
\bibliography{references}  

\end{document}